\documentclass[letterpaper]{article} 
\usepackage{aaai23}  
\usepackage{times}  
\usepackage{helvet}  
\usepackage{courier}  
\usepackage[hyphens]{url}  
\usepackage{graphicx} 
\usepackage{natbib}  
\usepackage{caption} 
\usepackage{algorithm}
\usepackage{algorithmic}

\usepackage{graphicx}
\usepackage{amsmath}
\usepackage{amssymb}
\usepackage{booktabs}
\usepackage{multirow}
\usepackage{ulem}
\usepackage{color}
\usepackage{pifont}
\usepackage{balance}
\usepackage{colortbl}
\usepackage{xcolor}
\usepackage{makecell}
\usepackage{amsmath,amssymb,mathrsfs}
\usepackage{bm}
\usepackage{algorithm, algorithmic}
\newcommand{\cmark}{\ding{51}}

\usepackage{balance}
\usepackage{bbding} 
\usepackage{multirow}
\definecolor{newcolor}{rgb}{.8,.349,.1}
\usepackage{newfloat}
\usepackage{listings}
\DeclareCaptionStyle{ruled}{labelfont=normalfont,labelsep=colon,strut=off} 
\floatstyle{ruled}
\newfloat{listing}{tb}{lst}{}
\floatname{listing}{Listing}
\nocopyright 

\title{URCDC-Depth: Uncertainty Rectified Cross-Distillation with \\  CutFlip for Monocular Depth Estimation}

\author {
	Shuwei Shao\textsuperscript{\rm 1},
	Zhongcai Pei\textsuperscript{\rm 1},
	Weihai Chen\textsuperscript{\rm 1},
	Ran Li\textsuperscript{\rm 1},
	Zhong Liu\textsuperscript{\rm 1},
	Zhengguo Li\textsuperscript{\rm 2},
	
}
\affiliations {
	\textsuperscript{\rm 1} School of Automation Science and Electrical Engineering, Beihang University\\
	\textsuperscript{\rm 2}Institute for Infocomm Research, A*STAR\\
	swshao@buaa.edu.cn\\
	
}

\usepackage{bibentry}

\begin{document}

\maketitle

\begin{abstract}
	This work aims to estimate a high-quality depth map from a single RGB image. Due to the lack of depth clues, making full use of the long-range correlation and the local information is critical for accurate depth estimation. Towards this end, we introduce an uncertainty rectified cross-distillation between Transformer and convolutional neural network (CNN) to learn a unified depth estimator. Specifically, we use the depth estimates from the Transformer branch and the CNN branch as pseudo labels to teach each other. Meanwhile, we model the pixel-wise depth uncertainty to rectify the loss weights of noisy pseudo labels. To avoid the large capacity gap induced by the strong Transformer branch deteriorating the cross-distillation, we transfer the feature maps from Transformer to CNN and design coupling units to assist the weak CNN branch to leverage the transferred features. Furthermore, we propose a surprisingly simple yet highly effective data augmentation technique CutFlip, which enforces the model to exploit more valuable clues apart from the vertical image position for depth inference. Extensive experiments demonstrate that our model, termed~\textbf{URCDC-Depth}, exceeds previous state-of-the-art methods on the KITTI, NYU-Depth-v2 and SUN RGB-D datasets, even with no additional computational burden at inference time. The source code is publicly available at~\url{https://github.com/ShuweiShao/URCDC-Depth}.
\end{abstract}

\section{Introduction}

Monocular depth estimation is a fundamental research topic in the computer vision community, with applications ranging from scene understanding, 3D reconstruction through to augmented reality. Benefiting from the advances in convolutional neural networks (CNNs)~\cite{He_2016_CVPR, tan2019efficientnet}, recent studies~\cite{lee2019big, bhat2021adabins} achieve promising depth results. 
Due to the lack of depth cues, fully exploiting the long-range correlation (i.e., inter-object distance relationship) and the local information (i.e., intra-object consistency), is crucial for accurate depth estimation~\cite{saxena2005learning}. However, the convolution operator with a limited receptive field is hard to capture the long-range correlation, which becomes a potential bottleneck of current CNN-based depth estimation methods~\cite{bhat2021adabins}.
\normalem
\begin{figure}[!htb]
	\centering
	\includegraphics[width=1.0\linewidth]{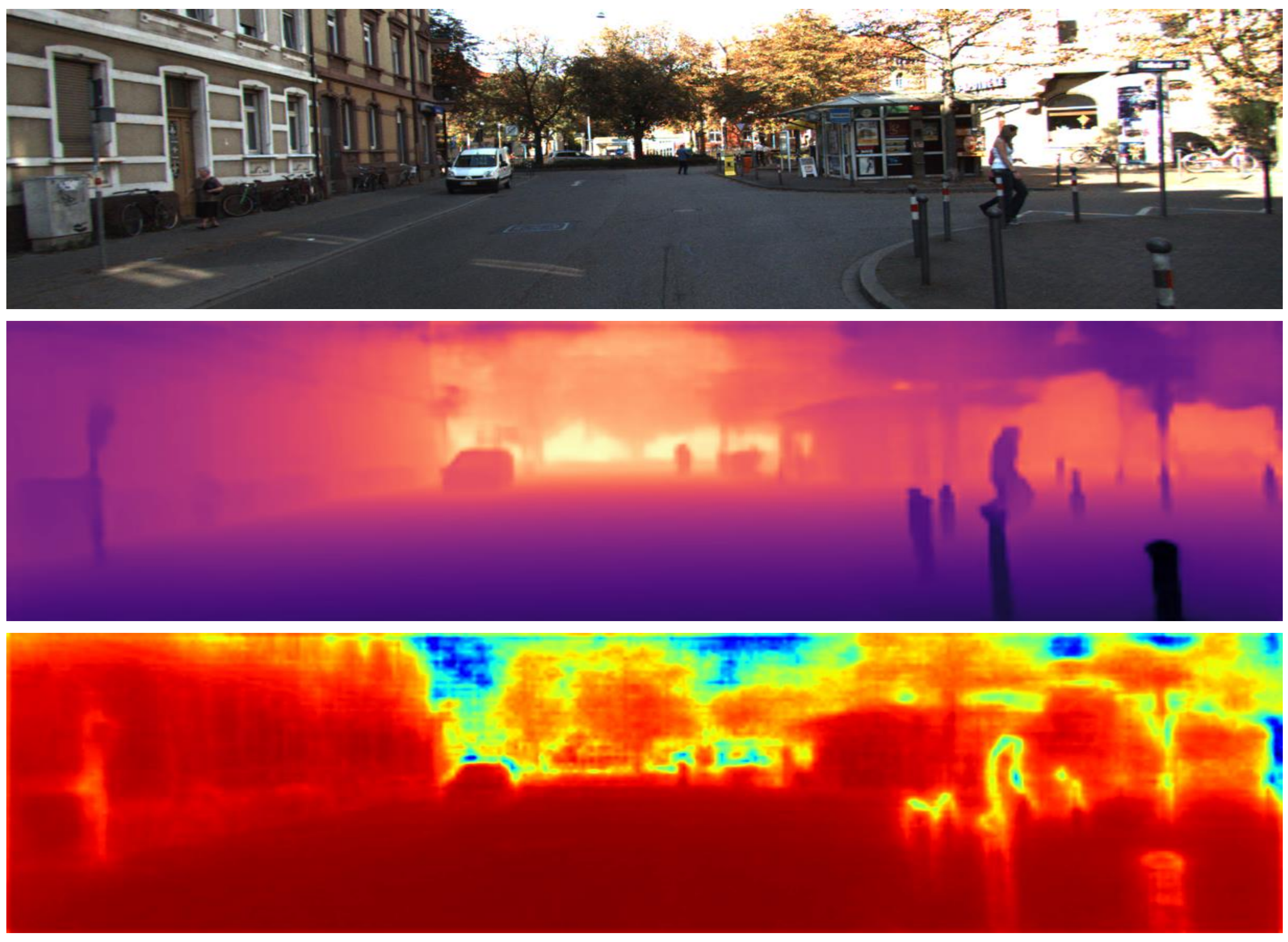}
	\caption{\textbf{Illustration of the depth and uncertainty maps from URCDC-Depth}. \emph{Top: input image; Middle: estimated depth map; Bottom: pixel-wise depth uncertainty (red: low uncertainty; yellow/blue: high/highest uncertainty).} }
	\label{Fig1}
\end{figure}

There are extensive works dedicated to alleviating the above limitation of CNN, which can be roughly divided into two categories: manipulating the convolution operation and introducing the attention mechanism~\cite{vaswani2017attention}. The former leverages atrous spatial pyramid pooling~\cite{chen2017deeplab}, coarse-to-fine fusion~\cite{lin2017feature} and densely connecting~\cite{zhang2020densely} to enhance the efficacy of convolution operator. The latter integrates the attention module to establish the long-distance dependency in the feature map~\cite{zhou2019unsupervised,johnston2020self}. In addition, several general methods adopt both of these strategies~\cite{huynh2020guiding, bhat2021adabins}. Despite the considerable improvements in performance, the dilemma remains.

\begin{figure*}[!htb]
	\centering
	\includegraphics[width=0.95\linewidth]{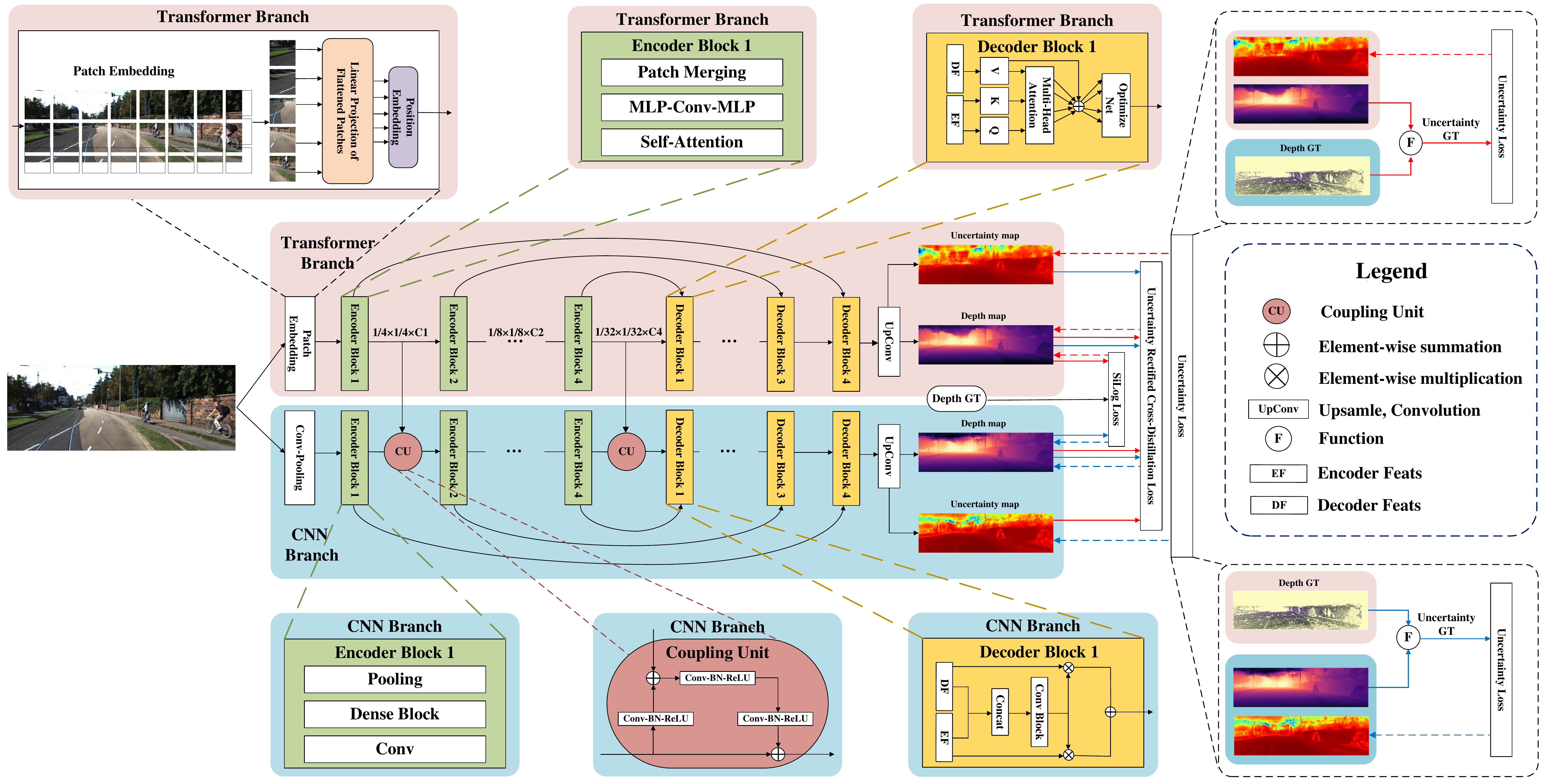}
	\caption{\textbf{Overview of the developed URCDC-Depth}. Our URCDC-Depth in the training phase consists of two branches, a Transformer branch and a CNN branch. During the evaluation phase, we only leverage the Transformer branch to generate the depth map.}
	\label{Fig2}
\end{figure*}

Recently, visual Transformer has been demonstrated as a promising alternative to the CNN~\cite{dosovitskiy2020image,Liu_2021_ICCV,Peng_2021_ICCV}. Building upon the attention mechanism, the Transformer with a global receptive field is more proficient in capturing the long-range correlation. Nevertheless, local feature details are prone to be ignored by it due to the lack of spatial inductive bias, resulting in unsatisfactory performance. A few depth estimation methods try to overcome the drawback of Transformer by utilizing additional CNN branch~\cite{li2022depthformer,shao2021nenet}. However, these frameworks also rely on the CNN branch in the evaluation phase, increasing the computational cost at inference time.

In this work, we introduce a novel monocular depth estimation framework, termed \textbf{URCDC-Depth} (Fig.~\ref{Fig2}), which integrates the strengths from both the Transformer and CNN using \textbf{cross-distillation} to enhance the performance. The core idea of URCDC-Depth lies in that the Transformer branch establishes the long-range correlation while the CNN branch focuses on the local information, so cross-distillation between these two branches can help learn a unified depth estimator with both properties. To be specific, we use the derived depth estimates as pseudo labels to teach their counterparts. To alleviate the negative impact of noisy labels, we model the \textbf{pixel-wise depth uncertainty} to rectify the loss weights in these regions. The uncertainty map is predicted jointly with the corresponding depth map. In addition, we transfer the feature maps from the Transformer to the CNN so as to bridge the large performance gap induced by the strong Transformer branch and design \textbf{coupling units} to assist the weak CNN branch to utilize the transferred feature maps, which contribute to boost the performance of cross-distillation.

Furthermore, we train the URCDC-Depth with a very simple yet effective data augmentation technique, \textbf{CutFlip}, based on the observation that monocular depth estimation model relies heavily on the vertical image position to infer depth, while other clues such as apparent sizes are ignored, deteriorating the model generalization ability~\cite{dijk2019neural}. Generally, the feature of vertical image position is in that the closer the projection on the image is to the lower boundary, the smaller the depth of the scene point. The traditional training mechanism allows the clue of vertical image position to exist in almost all training samples. In contrast, other cues are much less numerous. To resolve this, we vertically cut the training sample into upper and lower parts, and flip these two parts along the vertical direction with a certain probability, weakening the relationship between depth and vertical image position. In such case, the accuracy of predicted depth is significantly improved.

To summarize, the main contributions of this work are listed as:

\begin{itemize}
	\item We introduce a novel monocular depth estimation model equipped with  uncertainty rectified cross-distillation to exploit both the long-range correlation and the local information. Besides, the model has no additional computational burden in the evaluation phase thanks to the cross-distillation paradigm.
	\item We design a simple yet effective data augmentation strategy, which enforces the model to focus on more valuable cues for depth estimation, not just the clue of vertical image position.
	\item Detailed experiments and analysis indicate the efficacy of our developed components in improving the depth accuracy. The proposed approach achieves state-of-the-art performance on the KITTI~\cite{geiger2013vision} and NYU-Depth-v2~\cite{silberman2012indoor} datasets.
\end{itemize}


\section{Related work}
\textbf{Monocular depth estimation} attempts to regress depth map from a single RGB image. As a seminal work,~\citet{saxena2005learning} used a Markov random field to predict depth. Later, benefiting from the encoded features of CNNs that generalize well across diverse tasks, many follow-up works have achieved drastic performance improvement~\cite{eigen2014depth,qi2018geonet,fu2018deep}. Recently,~\citet{lee2019big} introduced local planar guidance layers to infer plane coefficients in the decoding stage, which were leveraged to recover the full resolution depth map.
~\citet{bhat2021adabins} revisited the ordinal regression network~\cite{fu2018deep} and proposed to calculate adaptive bins based on the image content.

\textbf{Transformer} has attracted a widespread attention owing to its effectiveness in natural language processing~\cite{vaswani2017attention}. In terms of computer vision,~\citet{dosovitskiy2020image} introduced Vision Transformer (ViT) and indicated its feasibility on the image classification task. The success of the ViT accelerates the application of the Transformer to other tasks.~\citet{zheng2021rethinking} developed one of the first attempts at dense prediction tasks by using the ViT as the backbone.

There have been some attempts at applying Transformer to monocular depth estimation.~\citet{bhat2021adabins} utilized a minimized version of ViT to calculate bin width adaptively.~\citet{yang2021transformer,ranftl2021vision,kim2022global,Yuan_2022_CVPR} used the Transformer as an encoder to attain a global receptive field. As demonstrated in~\cite{li2022depthformer}, the model with a Transformer encoder tends to lose local depth details, \textit{e.g.}, sharp edges.~\citet{li2022depthformer} proposed to use the Transformer encoder and an additional CNN encoder so that the model can enjoy the desired properties from both networks. However, the ensembled model increases the computational complexity. By contrast, we leverage the cross-distillation between the Transformer and CNN to construct a unified depth estimator, which allows our model to use only the Transformer branch at inference time with no additional computational burden.

\textbf{Knowledge distillation} is a learning paradigm targeting to transfer the learned knowledge from a teacher model to a lower-capacity student model, which is initially proposed on image recognition~\cite{hinton2015distilling}. Since then, numerous  knowledge distillation variants have been proposed, either working to improve its effectiveness~\cite{yim2017gift,tian2019contrastive,sun2019deeply} or applying it to other tasks~\cite{garcia2018modality,hafner2018cross}. Cross-distillation is a special case where models reach a consensus by simultaneously teaching each other, which is similar to the mutual learning~\cite{zhang2018deep}. A few studies also apply knowledge distillation to enhance monocular depth estimation~\cite{pilzer2019refine,aleotti2020real}. Unlike these methods, we introduce an uncertainty rectified cross-distillation for accurate depth estimation. 

\textbf{Data augmentation} is a powerful technique in mitigating overfitting by increasing the effective amount of training samples. Therefore, common data augmentation techniques such as color jitter, crop, rotation are used in various tasks to improve model performance. Besides, there are methods tailored for monocular depth estimation, CutDepth~\cite{ishii2021cutdepth} and DataGrafting~\cite{Peng_2021_ICCV}. The motivation of CutDepth differs significantly from our CutFlip. Concretely, the CutDepth aims to shorten the distance between the RGB image and the depth map in the latent space by replacing part of the RGB image with the corresponding depth ground-truth. While DataGrafting also aims to mitigate the overfitting risk for vertical image position, it relies on grafting together two training samples with different semantics. In contrast, our CutFlip is simpler and easier to implement as it only requires one training sample. 

\section{Methodology}
In this section, we elaborate on the main contributions of this work, namely uncertainty rectified cross-distillation and CutFlip. An overview of URCDC-Depth is presented in Fig.~\ref{Fig2}.
\subsection{Uncertainty Rectified Cross-Distillation}

\textbf{Network architecture}. The proposed Transformer branch shares a same network architecture with NeWCRFs~\cite{Yuan_2022_CVPR} apart from the final prediction layer, which generates not only the depth map, but also the pixel-wise depth uncertainty. The encoder uses the Swin Transformer~\cite{Liu_2021_ICCV} to extract hierarchical feature representations. The decoder is composed of four neural window fully-connected conditional random fields (CRFs) modules.

The proposed CNN branch is also based on an encoder-decoder structure, where the encoder is DenseNet~\cite{huang2017densely} and the decoder is similar to~\cite{kim2022global}. The CNN branch is only used for complementary training, and will be discarded once training process is complete.

\textbf{Cross-distillation between Transformer and CNN}. We make use of the cross-distillation to construct a unified depth estimator that fully exploits the long-range correlation and the local information. For an input RGB image $\mathbf{r}_n\left( \textbf{p} \right)$, where $\textbf{p}$ denotes the pixel coordinate, our model generates two depth predictions,
\begin{equation}
	\textbf{d}_n^t\left( \textbf{p} \right) = f_\theta ^t\left( {{\textbf{r}_n\left( \textbf{p} \right)}} \right);\textbf{d}_n^c\left( \textbf{p} \right) = f_\theta ^c\left( {{\textbf{r}_n\left( \textbf{p} \right)}} \right),
\end{equation}
where $\textbf{d}_n^t\left( \textbf{p} \right)$ and $\textbf{d}_n^c\left( \textbf{p} \right)$ denotes the predictions from Transformer branch $f_\theta ^t\left(  \cdot  \right)$ and CNN branch $f_\theta ^c\left(  \cdot  \right)$, respectively. As mentioned before, the Transformer and CNN are asymmetric learning networks, where the Transformer relies on the long-range self-attention while the CNN is built upon the local convolution operator, so the predictions $\textbf{d}_n^t\left( \textbf{p} \right)$ and $\textbf{d}_n^c\left( \textbf{p} \right)$ have inherently diverse properties and are used as pseudo labels to guide their counterparts towards the correct depth. 

\textbf{Uncertainty-based rectification.} However, the pseudo labels contain heavy noises, particularly at the beginning of training, which inevitably damage the entire training process and enforce wrong predictions. To alleviate the negative impact of depth noises, we introduce an uncertainty rectified cross-distillation loss, defined as
\begin{equation}
	\begin{split}
		{\mathcal{L}_{urcd}} = \sum\limits_\textbf{p} \left( {1 - \overline {\textbf{u}_n^c} }\left( \textbf{p} \right) \right) \odot {\left| {\textbf{d}_n^t\left( \textbf{p} \right) - \overline {\textbf{d}_n^c} \left( \textbf{p} \right)} \right|}\\ + \sum\limits_\textbf{p} \left( {1 - \overline {\textbf{u}_n^t} }\left( \textbf{p} \right) \right)\odot {\left| {\textbf{d}_n^c\left( \textbf{p} \right) - \overline {\textbf{d}_n^t} \left( \textbf{p} \right)} \right|},
	\end{split}
\end{equation}
\begin{figure}[!htb]
	\centering
	\includegraphics[width=0.9\linewidth]{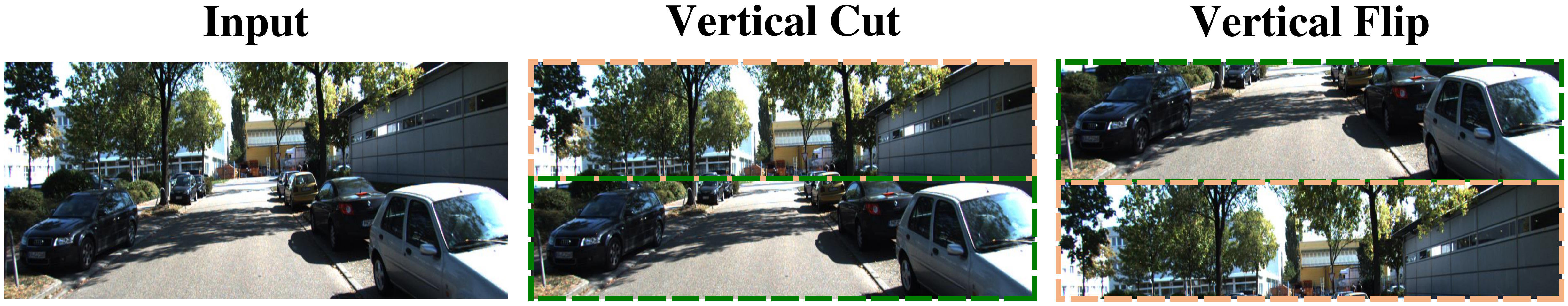}
	\caption{\textbf{Illustration of the technique CutFlip}. The CutFilp contains two key steps: vertical cut and vertical filp. }
	\label{Fig3}
\end{figure}
where $\overline \cdot$ is the gradient stopping operation, $\odot$ is the element-wise multiplication, and $\textbf{u}_n^t\left( \textbf{p} \right)$ and $\textbf{u}_n^c\left( \textbf{p} \right)$ are the uncertainty predictions from Transformer branch $f_\theta ^t\left(  \cdot  \right)$ and CNN branch $f_\theta ^c\left(  \cdot  \right)$, respectively, which are used to downweight the relevant pixels to alleviate the negative impact on regions with high uncertainty. The values of uncertainty map are ranging from 0 to 1.

Since there is no ground-truth for the uncertainty prediction, we model it with a function inspired by the probability density function of Laplace distribution, mathematically,
\begin{equation}
	\textbf{u}_n^*\left( \textbf{p} \right) = 1 - \exp \left( { - \frac{{\left| {{\textbf{d}_n}\left( \textbf{p} \right) - \textbf{d}_n^*\left( \textbf{p} \right)} \right|}}{b\left( {{\textbf{d}_n}\left( \textbf{p} \right) + \textbf{d}_n^*\left( \textbf{p} \right)} \right) }} \right),\textbf{p} \in \textbf{T},
\end{equation}
where ${\textbf{d}_n}\left( \textbf{p} \right)$ stands for the predicted depth map, $\textbf{d}_n^*\left( \textbf{p} \right)$ denotes the ground-truth depth map, $b$ is a coefficient that controls the tolerance for error and is set as 0.2 in this work, and \textbf{T} denotes a set of pixels with valid ground-truth depth values. Here, instead of using the absolute difference between ${\textbf{d}_n}\left( \textbf{p} \right)$ and $\textbf{d}_n^*\left( \textbf{p} \right)$ directly, we normalize it by their sum due to the fact that a unit depth difference (\textit{e.g.}, 1m) represents the different uncertainty between distant and nearby points in a scene, and it should be higher on nearby points and less on distant points. We apply $\mathcal{L}_u$ to enfore the uncertainty prediction to approximate $\textbf{u}_n^*\left( \textbf{p} \right)$,
\begin{equation}	\resizebox{0.90\hsize}{!}{${\mathcal{L}_{u}} = \sum\limits_\textbf{p} {\left| {\textbf{u}_n^t\left( \textbf{p} \right) - {\textbf{u}_n^{t*}} \left( \textbf{p} \right)} \right|}  + \sum\limits_\textbf{p} {\left| {\textbf{u}_n^c\left( \textbf{p} \right) - {\textbf{u}_n^{c*}} \left( \textbf{p} \right)} \right|}, \textbf{p} \in \textbf{T}.$}
\end{equation}

\textbf{Coupling unit}. As presented in recent studies~\cite{li2022depthformer, Yuan_2022_CVPR, lee2019big}, the Swin Transformer-based models perform much better than CNN-based ones for depth estimation. Moreover, the large capacity gap between teacher and student tends to cause poor performance of knowledge distillation ~\cite{hu2021boosting}. To bridge it, we transfer the feature maps encoded by the Transformer branch to the CNN branch, and design \textbf{coupling units} to fuse these two types of features. First, we align the channel dimension of the feature maps via a $1 \times 1$ convolution and add them together. Second, we use a 3x3 convolution to adaptively fuse the added features. Finally, we adjust the feature channel dimension with a $1 \times 1$ convolution to form a residual connection. Meanwhile, BatchNorm~\cite{ioffe2015batch} and ReLU activation are utilized to regularize features. Note that the feature transfer operation only exists in the encoder that mainly determines the model performance.

\begin{algorithm}
	\renewcommand{\algorithmicrequire}{\textbf{Input:}}
	\renewcommand{\algorithmicensure}{\textbf{Output:}}
	\caption{CutFlip}
	\label{alg1}
	\begin{algorithmic}[1]
		\REQUIRE RGB image $\textbf{r}_n$; Ground-truth depth map $\textbf{d}_n^*$; Height of input $h$.
		
		\ENSURE Transformed $\textbf{r}_n$; Transformed $\textbf{d}_n^*$.
		\STATE Random sampling $p$ from the uniform distribution \textbf{U} (0, 1).
		\IF {$p<0.5$}
		\RETURN $\textbf{r}_n$; $\textbf{d}_n^*$.
		\ELSE
		\STATE Random sampling $\varsigma$ from [floor($0.2h$), floor($0.8h$)];
		\STATE $\textbf{r}={\textbf{r}_n}.copy()$; $\textbf{d}^*={\textbf{d}_n^*}.copy()$;
		\STATE $\textbf{r}_n[:h-\varsigma, :, :]$ = $\textbf{r}[\varsigma:, :, :]$; $\textbf{d}_n^*[:h-\varsigma, :, :]$ = $\textbf{d}^*[\varsigma:, :, :]$;
		\STATE $\textbf{r}_n[h-\varsigma:, :, :]$ = $\textbf{r}[:\varsigma, :, :]$; $\textbf{d}_n^*[h-\varsigma:, :, :]$ = $\textbf{d}^*[:\varsigma, :, :]$;
		\RETURN $\textbf{r}_n$; $\textbf{d}_n^*$.
		\ENDIF
	\end{algorithmic}
\end{algorithm}
\textbf{Overall loss}. The total optimization objective to train the URCDC-Depth is summarized as follows
\begin{equation} {\mathcal{L}_{total}} = {\mathcal{L}_{ssi}} + {\lambda _1}{\mathcal{L}_{urcd}} + {\lambda _2}{\mathcal{L}_u},
\end{equation}
with
\begin{equation}
	\begin{array}{l}
		{\mathcal{L}_{ssi}} = \kappa \sqrt {\frac{1}{{\left| \textbf{T} \right|}}\sum\limits_\textbf{p} {{{\left( {\textbf{g}_n^t\left( \textbf{p} \right)} \right)}^2} - \frac{\eta }{{{{\left| \textbf{T} \right|}^2}}} {{\left( {\sum\limits_\textbf{p} {\textbf{g}_n^t\left( \textbf{p} \right)} } \right)}^2}} } \\
		+ \kappa \sqrt {\frac{1}{{\left| \textbf{T} \right|}} \sum\limits_\textbf{p} {{{\left( {\textbf{g}_n^c\left( \textbf{p} \right)} \right)}^2} - \frac{\eta }{{{{\left| \textbf{T} \right|}^2}}} {{\left( {\sum\limits_\textbf{p} {\textbf{g}_n^c\left( \textbf{p} \right)} } \right)}^2}}}, \textbf{p} \in \textbf{T},
	\end{array}
\end{equation}
where ${\mathcal{L}_{ssi}}$ denotes the scaled scale-invariant loss introduced by~\cite{lee2019big}, ${\textbf{g}_n}\left( \textbf{p} \right) = \log {\textbf{d}_n}\left( \textbf{p} \right) - \log \textbf{d}_n^*\left( \textbf{p} \right)$, $\kappa$ and $\eta$ are set as 10 and 0.85 based on~\cite{lee2019big}, $\lambda _1$ and $\lambda _2$ are empirically set as 0.1 and 0.5.

\subsection{CutFlip}

The quantity and diversity of training data are critical for deep learning-based models, which however are hard to satisfy in depth estimation because data acquisition is extremely expensive and laborious. Lack of data deteriorates the model generalization ability, and one of the serious overfitting threats is the heavy reliance on the vertical image position~\cite{dijk2019neural}. To enforce the model to focus on more valuable clues, we propose a surprisingly simple yet highly effective data augmentation technique, \textbf{CutFlip}. As illustrated in Fig.~\ref{Fig3}, we vertically cut the input sample into upper and lower parts, highlighted by the orange box and the green box, respectively, and flip these two parts along the vertical direction to weaken the relationship of depth and vertical image position. The details are shown in Algorithm~\ref{alg1}. The CutFlip is performed with a probability of 0.5, and the vertical position to cut is randomly sampled, which allows the model to greatly adapt to various types of data.

\section{Experiments}
\begin{figure*}[!]
	\centering
	\includegraphics[width=0.95\linewidth]{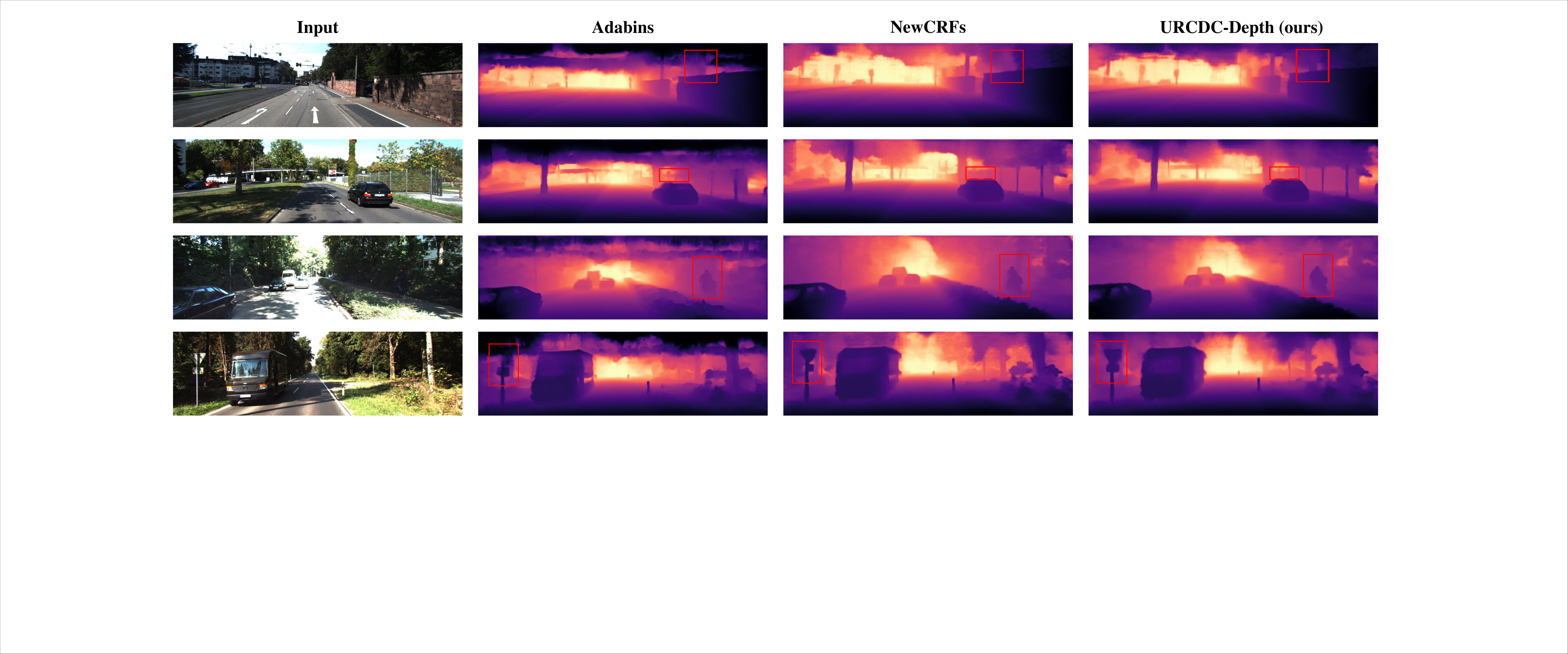}
	\caption{\textbf{Qualitative depth results on the Eigen split of KITTI dataset}. The red boxes indicate the regions to emphasize.}
	\label{Fig5}
\end{figure*}
\begin{table*}[htb!]
	\begin{center}
		\renewcommand{\arraystretch}{1.3}
		\resizebox{2.0\columnwidth}{!}{\begin{tabular}{c|| c || c c  c c || c c c }	
				\Xhline{1.2pt}
				Method & Cap & Abs Rel $\downarrow$ & Sq Rel $\downarrow$ & RMSE $\downarrow$ & RMSE log $\downarrow$ & $\delta  < 1.25$ $\uparrow$ & $\delta  < {1.25^2}$ $\uparrow$& $\delta  < {1.25^3}$ $\uparrow$ \\
				\hline						
				\hline
				Eigen~\textit{et al.}~\cite{eigen2015predicting}&0-80m&0.203&1.548&6.307&0.282&0.702&0.898&0.967
				\\
				Fu~\textit{et al.}~\cite{fu2018deep}&0-80m&0.072&0.307&2.727&0.120&0.932&0.984&0.994
				\\
				VNL~\cite{yin2019enforcing}&0-80m&0.072&-&3.258&0.117&0.938&0.990&0.998
				\\
				BTS~\cite{lee2019big}&0-80m&0.061&0.261&2.834&0.099&0.954&0.992&0.998
				\\
				PWA~\cite{lee2021patch}&0-80m&0.060&0.221&2.604&0.093&0.958&0.994&\textbf{0.999}
				\\
				TransDepth~\cite{yang2021transformer}&0-80m&0.064&0.252&2.755&0.098&0.956&0.994&\textbf{0.999}
				\\
				Adabins~\cite{bhat2021adabins}&0-80m&0.058&0.190&2.360&0.088&0.964&0.995&\textbf{0.999}
				\\ 		
				P3Depth~\cite{patil2022p3depth}&0-80m&0.071&0.270&2.842&0.103&0.953&0.993&0.998
				\\
				DepthFormer~\cite{li2022depthformer}&0-80m&0.052&0.158&2.143&0.079&0.975&\textbf{0.997}&\textbf{0.999}
				\\
				NeWCRFs~\cite{Yuan_2022_CVPR}&0-80m&0.052&0.155&2.129&0.079&0.974&\textbf{0.997}&\textbf{0.999}
				\\				
				\hline
				\textbf{URCDC-Depth (ours)} &0-80m &\textbf{0.050}&\textbf{0.142}&\textbf{2.032}&\textbf{0.076}&\textbf{0.977}&\textbf{0.997}&\textbf{0.999}\\
				\hline
				\hline
				Fu~\textit{et al.}~\cite{fu2018deep}&0-50m&0.071&0.268&2.271&0.116&0.936&0.985&0.995
				\\
				BTS~\cite{lee2019big}&0-50m&0.058&0.183&1.995&0.090&0.962&0.994&0.999
				\\
				PWA~\cite{lee2021patch}&0-50m&0.057&0.161&1.872&0.087&0.965&0.995&0.999
				\\
				TransDepth~\cite{yang2021transformer}&0-50m&0.061&0.185&1.992&0.091&0.963&0.995&0.999
				\\
				P3Depth~\cite{patil2022p3depth}&0-50m&0.055&0.130&1.651&0.081&0.974&0.997&0.999
				\\
				\hline
				\textbf{URCDC-Depth (ours)} &0-50m&\textbf{0.049}&\textbf{0.108}&\textbf{1.528}&\textbf{0.072}&\textbf{0.981}&\textbf{0.998} &\textbf{1.000}\\
				\Xhline{1.2pt}
				
		\end{tabular}}
	\end{center}
	\caption{\textbf{Quantitative depth comparison on the Eigen split of KITTI dataset}. Note that the backbones of DepthFormer, NeWCRFs and our URCDC-Depth at inference time are Swin-Large and ResNet-50, Swin-Large and Swin-Large, respectively. ``-'' indicates not applicable. The best results are highlighted in \textbf{bold}.}
	\label{table2}
\end{table*}

\begin{table}[htb!]
	\begin{center}
		\renewcommand{\arraystretch}{1.3}
		\resizebox{1.0\columnwidth}{!}{\begin{tabular}{c|| c c c c  }	
				\Xhline{1.2pt}
				Method & SILog $\downarrow$ & sqErrRel $\downarrow$ & absErrRel $\downarrow$ & iRMSE$\downarrow$\\
				\hline						
				\hline
				Fu~\textit{et al.}~\cite{fu2018deep}&11.77&8.78&2.23&12.98
				\\
				BTS~\cite{lee2019big}&11.67&9.04&2.21&12.23
				\\
				BA-Full~\cite{aich2021bidirectional}&11.61&9.38&2.29&12.23
				\\
				PackNet-SAN~\cite{guizilini2021sparse}&11.54&9.12&2.35&12.38
				\\
				PWA~\cite{lee2021patch}&11.45&9.05&2.30&12.32
				\\
				NeWCRFs~\cite{Yuan_2022_CVPR}&10.39&8.37&1.83&11.03
				\\
				\hline
				\textbf{URCDC-Depth (ours)} &\textbf{10.03} &\textbf{8.24}&\textbf{1.74}&\textbf{10.71}
				\\
				\Xhline{1.2pt}		
		\end{tabular}}
	\end{center}
	\caption{\textbf{Quantitative depth comparison on the official split of KITTI dataset}. The results are available from the online server.} 
	\label{table3}
\end{table}
\subsection{Datasets}
\textbf{KITTI dataset} is captured from outdoor scenes with equipment placed on a moving vehicle~\cite{geiger2013vision}. The image resolution is around $1241 \times 376$ pixels. We use two commonly used splits for monocular depth estimation. One is the Eigen split~\cite{eigen2014depth} including 23488 training image pairs and 697 testing images. The other one is the official split~\cite{geiger2013vision} including 42949 training image pairs, 1000 validation images and 500 testing images. The evaluation results on the official split are generated by the online server.

\textbf{NYU-Depth-v2 dataset} provides RGB images and depth maps collected from indoor scenes at a resolution of $640 \times 480$ pixels~\cite{silberman2012indoor}. Following prior works, we adopt the official split and the dataset processed by~\citet{lee2019big}, which contains 24231 training images and 654 testing images.

\begin{figure*}[!]
	\centering
	\includegraphics[width=0.92\linewidth]{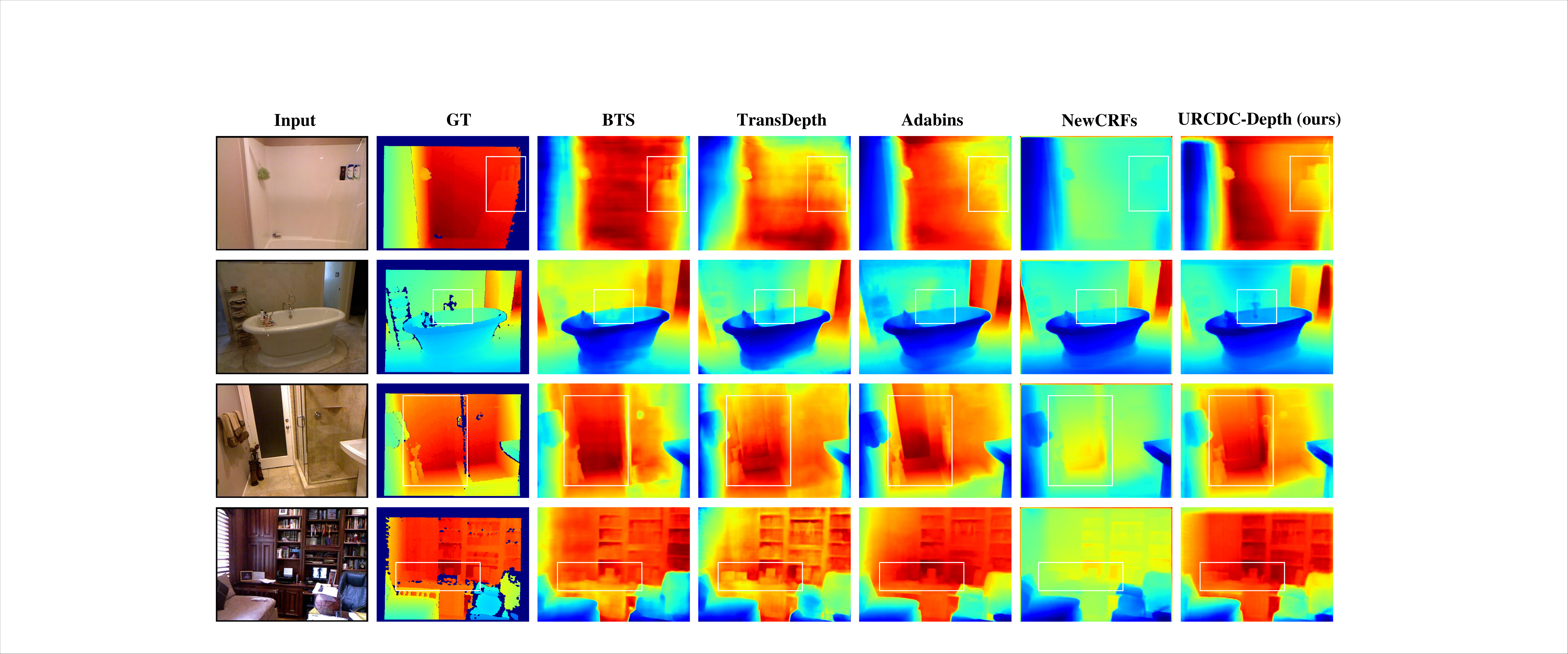}
	\caption{\textbf{Qualitative depth results on the NYU-Depth-v2 dataset}. The white boxes indicate the regions to emphasize.}
	\label{Fig6}
\end{figure*}
\begin{table*}[htb!]
	\begin{center}
		\renewcommand{\arraystretch}{1.3}
		\resizebox{2.0\columnwidth}{!}{\begin{tabular}{c || c || c c c || c c c}	
				\Xhline{1.2pt}
				Method &  Cap & Abs Rel $\downarrow$ & RMSE $\downarrow$ & ${\textbf{\rm{log}}_{\bm{{10}}}}$ $\downarrow$ &  $\delta  < 1.25$ $\uparrow$ &  $\delta  < {1.25^2}$ $\uparrow$& $\delta  < {1.25^3}$ $\uparrow$ \\
				\hline						
				\hline
				Eigen~\textit{et al.}~\cite{eigen2014depth}& 0-10m& 0.158&0.641&-&0.769&0.950&0.988
				\\
				Fu~\textit{et al.}~\cite{fu2018deep}& 0-10m&0.115&0.509&0.051&0.828&0.965&0.992
				\\
				VNL~\cite{yin2019enforcing}& 0-10m&0.108&0.416&0.048&0.875&0.976&0.994
				\\
				BTS~\cite{lee2019big}& 0-10m&0.113&0.407&0.049&0.871&0.977&0.995
				\\
				DAV~\cite{huynh2020guiding}& 0-10m&0.108&0.412&-&0.882&0.980&0.996
				\\
				PWA~\cite{lee2021patch}& 0-10m&0.105&0.374&0.045&0.892&0.985&0.997
				\\
				Long~\textit{et al.}~\cite{Long_2021_ICCV}& 0-10m&0.101&0.377&0.044&0.890&0.982&0.996
				\\
				TransDepth~\cite{yang2021transformer}& 0-10m&0.106&0.365&0.045&0.900&0.983&0.996
				\\
				Adabins~\cite{bhat2021adabins}& 0-10m&0.103&0.364&0.044&0.903&0.984&0.997
				\\
				P3Depth~\cite{patil2022p3depth}& 0-10m&0.104&0.356&0.043&0.898&0.981&0.996
				\\ 	
				DepthFormer~\cite{li2022depthformer}& 0-10m&0.096&0.339&0.041&0.921&0.989&0.998
				\\ 	
				NeWCRFs~\cite{patil2022p3depth}& 0-10m&0.095&0.334&0.041&0.922&\textbf{0.992}&\textbf{0.998}
				\\ 					
				\hline
				\textbf{URCDC-Depth (ours)} & 0-10m&\textbf{0.088} &\textbf{0.316}&\textbf{0.038}&\textbf{0.933}&\textbf{0.992}&\textbf{0.998}\\
				\Xhline{1.2pt}
		\end{tabular}}
	\end{center}
	\caption{\textbf{Quantitative depth comparison on the NYU-Depth-v2 dataset}. } 
	\label{table1}
\end{table*}
\subsection{Implementation Details}
The URCDC-Depth is implemented in PyTorch~\cite{paszke2017automatic} and trained on NVIDIA RTX A5000 GPUs. We optimize it using the Adam optimizer~\cite{kingma2015adam} where $\beta 1=0.9, \beta 2=0.999$. The training process runs a total number of 20 epochs with a batch size of 8 and a learning rate scheduled via polynomial decay from 1e-4 to 1e-5. We use the standard data augmentation techniques and evaluation metrics following previous works~\cite{Yuan_2022_CVPR, lee2019big}.

\subsection{Comparison to State-of-the-Arts}
\textbf{KITTI.} We first conduct comparison with the leading methods on the Eigen split. Table~\ref{table2} shows the results, indicating that our URCDC-Depth exceeds previous methods. It is worth noting that although URCDC-Depth and NewCRFs share the almost identical network structure in the evaluation phase, it improves the NeWCRFs by 8.4$\%$ and 4.6$\%$ on the Sq Rel and RMSE. Fig.~\ref{Fig5} presents qualitative depth comparisons. As we can see, the NeWCRFs is struggle with thinner structures,~\textit{e.g.}, posts and difficult object boundaries such as human boundary, while our URCDC-Depth is capable of estimating these small details, which supports our standpoint that the cross-distillation helps to learn a unified depth estimator with both desired properties from the Transformer and CNN.

We then compare our URCDC-Depth against the competing methods on the official split. The results are generated by the online server and reported in Table~\ref{table3}. Here we can see that our URCDC-Depth outperforms previous methods again. A notable phenomenon is that the main ranking metric SILog, from the model proposed by Fu et al. to the PWA, has only increased by 2.5$\%$ in three years. With the advent of visual Transformer, the NeWCRFs makes a significant breakthrough on this metric via the designed neural CRFs. Our URCDC-Depth further improves the NeWCRFs by 3.5$\%$ on the SILog, even with no additional computational burden at inference time. 
\begin{table}[htb!]
	\begin{center}
		\renewcommand{\arraystretch}{1.3}
		\resizebox{1.0\columnwidth}{!}{\begin{tabular}{c c c c c c c c c}
				\Xhline{1.2pt}
				ID & CD & UP & CU & CF& Abs Rel $\downarrow$ & Sq Rel $\downarrow$ & RMSE $\downarrow$ & $\delta  < 1.25$ $\uparrow$\\
				\hline
				\hline				
				1&&& && 0.052 & 0.155 & 2.129 & 0.974\\
				2 &\cmark&&&& 0.055 & 0.155 & 2.086 & 0.975\\	
				3 &\cmark&\cmark&&& 0.051 & 0.147 & 2.076 & 0.976\\	
				4 &\cmark&&\cmark&& 0.051 & 0.148 & 2.078 & 0.975\\
				5 &\cmark&\cmark&\cmark&& 0.051& 0.147 & 2.062 & 0.976 \\
				6 &&&&\cmark& 0.051& 0.144 & 2.056 & \textbf{0.977} \\
				7 &\cmark&\cmark&\cmark&\cmark& \textbf{0.050}& \textbf{0.142} & \textbf{2.032} &\textbf{0.977} \\
				\Xhline{1.2pt}			
		\end{tabular}}
	\end{center}
	\caption{\textbf{Ablation study of the proposed URCDC-Depth on the KITTI dataset}. 
		CD: cross-distillation; UP: uncertainty map; CU: coupling unit; CF: CutFlip. }
	\label{table5}
\end{table}

\begin{table}[htb!]
	\begin{center}
		\renewcommand{\arraystretch}{1.3}
		\resizebox{1.0\columnwidth}{!}{\begin{tabular}{c c c c c c c c c}
				\Xhline{1.2pt}
				ID & CD & UP & CU & CF& Abs Rel $\downarrow$ & RMSE $\downarrow$ &  ${\textbf{\rm{log}}_{\bm{{10}}}}$ $\downarrow$ & $\delta  < 1.25$ $\uparrow$\\
				\hline
				\hline				
				1&&& && 0.095 & 0.334 & 0.041 & 0.922\\
				2 &\cmark&&&& 0.095 & 0.329 & 0.040 & 0.923\\	
				3 &\cmark&\cmark&&& 0.091 & 0.323 & 0.039 & 0.928\\	
				4 &\cmark&&\cmark&& 0.091 & 0.326 & 0.039 & 0.926\\
				5 &\cmark&\cmark&\cmark&& 0.089& 0.319 & \textbf{0.038} & 0.931 \\
				6 &&&&\cmark& 0.091& 0.322 & 0.039 & \textbf{0.934} \\
				7 &\cmark&\cmark&\cmark&\cmark& \textbf{0.088}& \textbf{0.316} & \textbf{0.038} & 0.933 \\
				\Xhline{1.2pt}			
		\end{tabular}}
	\end{center}
	\caption{\textbf{Ablation study of the proposed URCDC-Depth on the NYU-Depth-v2 dataset}. }
	\label{table4}
\end{table}

\begin{table}[htb!]
	\begin{center}
		\renewcommand{\arraystretch}{1.3}
		\resizebox{1.0\columnwidth}{!}{\begin{tabular}{c|| c c c c  }	
				\Xhline{1.2pt}
				Augmentation method& Abs Rel $\downarrow$ & Sq Rel $\downarrow$ & RMSE $\downarrow$ & $\delta  < 1.25$ $\uparrow$\\
				\hline						
				\hline
				CutMix~\cite{yun2019cutmix}&0.054&0.154&2.093&0.974
				\\
				CutOut~\cite{devries2017improved}&0.052&0.148&2.078&0.975
				\\
				CutDepth~\cite{ishii2021cutdepth}&0.052&0.150&2.076&0.975
				\\
				DataGrafting~\cite{Peng_2021_ICCV}&0.051&0.146&2.051&0.976
				\\
				\hline
				\textbf{CutFlip} &\textbf{0.050} &\textbf{0.142}&\textbf{2.032}&\textbf{0.977}
				\\
				\Xhline{1.2pt}		
		\end{tabular}}
	\end{center}
	\caption{\textbf{Comparison of data augmentation techniques on the KITTI dataset}. 
	}
	\label{table6}
\end{table}
\textbf{NYU-Depth-v2.} To demonstrate the competitiveness of our URCDC-Depth in the indoor scenario, we also evaluate it on the NYU-Depth-v2 dataset. The results are reported in Table~\ref{table1}, which indicates that our method greatly boosts the performance on most metrics, such as Abs Rel and RMSE. This emphasizes our  contributions in improving the results. We display qualitative depth comparisons in Fig.~\ref{Fig6}. As can be seen, our URCDC-Depth preserves small details~\textit{e.g.}, handle and predicts sharp depth edges even in scenes with extremely scarce texture (top row).

\begin{figure}[!]
	\centering
	\includegraphics[width=1.0\linewidth]{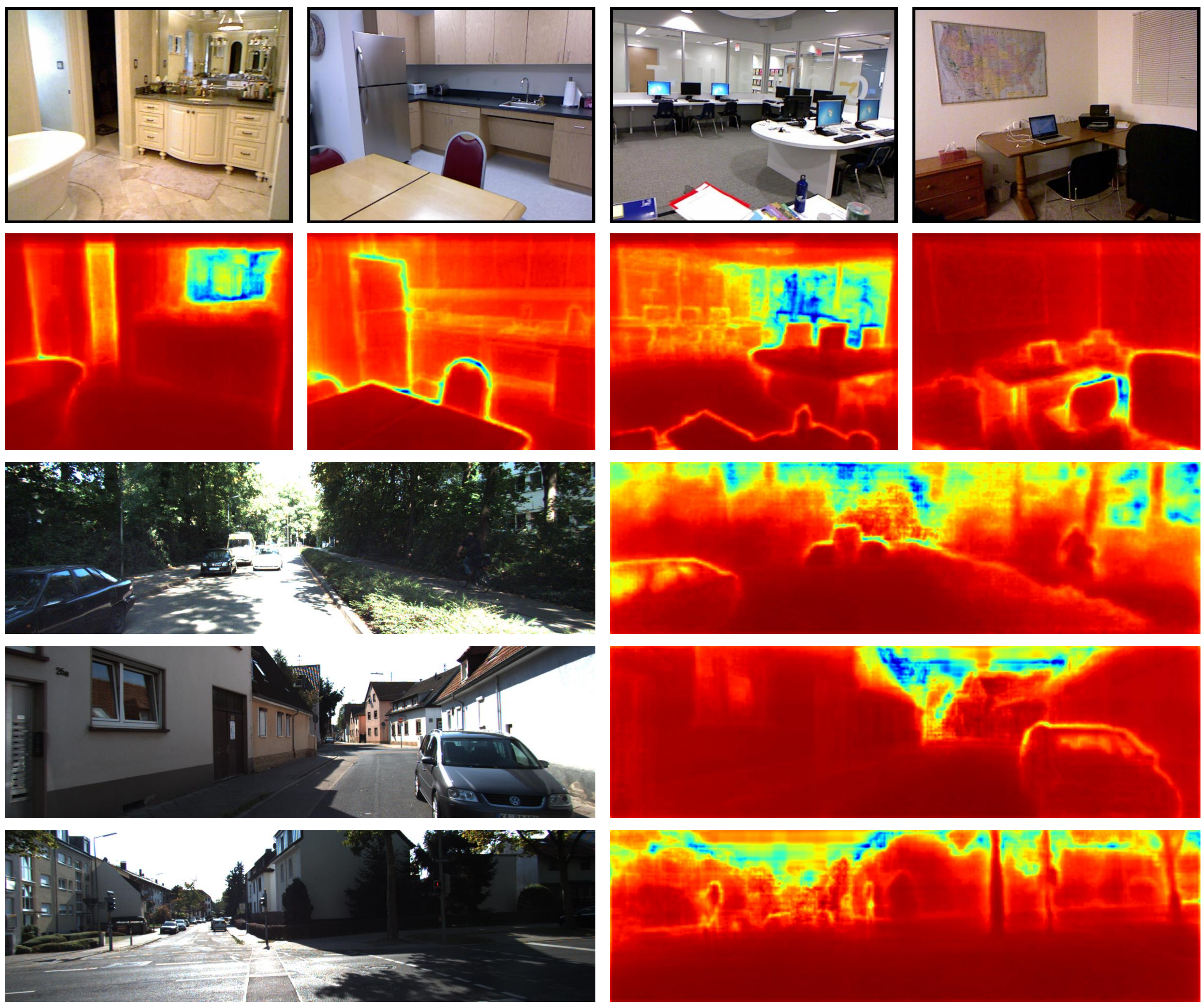}
	\caption{\textbf{Pixel-wise depth uncertainty from our URCDC-Depth}. Red indicates areas of low uncertainty, yellow/blue indicates areas of high/highest uncertainty.}
	\label{Fig7}
\end{figure}

\subsection{Ablation Study}
To better inspect how the proposed components in URCDC-Depth affect the performance, we present detailed ablation studies on the KITTI and NYU-Depth-v2 datasets, which are shown in Table~\ref{table5} and Table~\ref{table4}, respectively.

\textbf{Cross-distillation.} We start from the baseline NeWCRFs (ID 1). By directly introducing the cross-distillation, we observe a slight performance improvement on most metrics (ID 2). The cross-distillation suffers from the heavy depth noises from pseudo labels. Besides, the large capacity gap between Transformer branch and CNN branch limits the performance gain from the cross-distillation.

\textbf{Uncertainty map.} ID 3 presents the addition of the uncertainty map, which greatly boosts the performance. Especially notable are the results on Sq Rel sensitive to the large depth errors in Table~\ref{table5}. The uncertainty map helps mitigate the negative impact of depth noises in the training process, hence resulting in a considerably lower Sq Rel result. Fig.~\ref{Fig7} presents a visualization of uncertainty maps. Unlike the prior method~\cite{johnston2020self} that is also capable of estimating uncertainty maps showing a clear trend where uncertainty increases with distance, our uncertainty maps focus on the difficult regions, such as object boundaries and vanishing points. We attribute this to the normalization operation when modeling the ground-truth of uncertainty map.

\textbf{Coupling unit.} To bridge the large performance gap between the Transformer branch and CNN branch, we transfer the feature maps and use the coupling units to fuse the transferred features from the Transformer branch and the features in the CNN branch. The results after adding the coupling units are in IDs 4 and 5, which contributes to the performance.

\textbf{CutFlip.} With the CutFlip data augmentation, we can see that the performance is improved significantly (IDs 6 and 7). Besides, we compare the CutFlip against other similar augmentation methods to further demonstrate its efficacy in Table 6. To make a fair comparison, we remain all other configurations the same except for these specially crafted data augmentation techniques. The inferior performance of CutMix, CutOut and CutDepth may lie in the lack of constraint on the vertical image position. DataGrafting takes the overfitting risk of vertical image position into account. However, grafting together two training samples with different semantics increases the learning burden of network.

\section{Conclusion}
In this work, we introduce a novel monocular depth estimation framework URCDC-Depth, which leverages uncertainty rectified cross-distillation to fully exploit the long-range correlation and the local information. The paradigm allows our framework to generate precisely estimated depth maps with no additional computational burden at inference time. In addition, we propose a simple yet effective data augmentation technique CutFlip to enforce the model to emphasize more valuable depth reasoning clues apart from the vertical image position. We conduct comprehensive experiments on the KITTI, NYU-Depth-v2 and SUN RGB-D datasets, and the experimental results verify the efficacy of the proposed URCDC-Depth.

\normalem
\bibliography{reference}

\end{document}